\documentclass[letterpaper, 10 pt, journal, twoside]{ieeeconf}

\usepackage{graphics}    
\usepackage{times}       
\usepackage{amsmath}     
\usepackage{amssymb}     
\usepackage{graphicx}
\usepackage{algorithm}
\usepackage[noend]{algpseudocode}

\def\figref#1{Fig.~\ref{#1}}
\def\tabref#1{Tab.~\ref{#1}}
\def\eqref#1{Eq.~(\ref{#1})}

\usepackage{mathtools}

\DeclarePairedDelimiter{\sqnorm}{\lVert}{\rVert_2^2}

\makeatletter
\newcommand\footnoteref[1]{\protected@xdef\@thefnmark{\ref{#1}}\@footnotemark}
\makeatother

\usepackage{subcaption}

\usepackage[dvipsnames,table]{xcolor} 
\usepackage{hhline}

\usepackage{hyperref}
\hypersetup{
	colorlinks=true,
	linkcolor=red,
	filecolor=black,      
	urlcolor=blue,
}
\usepackage{pifont}
\usepackage{etoolbox}
\makeatletter
\patchcmd{\@makecaption}
{\scshape}
{}
{}
{}
\makeatletter
\patchcmd{\@makecaption}
{\\}
{:\ }
{}
{}
\makeatother

\usepackage[inline]{enumitem}

\usepackage{multirow}

\title{Learning to See the Wood for the Trees: \\Deep Laser
	Localization in Urban and Natural Environments on a CPU}

\author{Georgi Tinchev, Adrian Penate-Sanchez, and Maurice Fallon%
	\thanks{Manuscript received: September, 10, 2018; Revised December, 5, 
	2018; Accepted January, 8, 2019.}%
	\thanks{This paper was recommended for publication by Editor Cyrill Stachniss 
	upon evaluation of the Associate Editor and Reviewers' comments. 
		This work was supported by EPSRC RAIN and ORCA Robotics Hubs 
		(EP/R026084/1 and EP/R026173/1 respectively). M. Fallon is supported by a 
		Royal Society University Research Fellowship.}%
	\thanks{The authors are with the Dynamic Systems Group, Oxford Robotics 
	Institute, University of Oxford, United Kingdom. 
		\texttt{\{gtinchev,adrian,mfallon\}@robots.ox.ac.uk}}
	\thanks{Digital Object Identifier (DOI): see top of this page.}
}

\begin{document}

\markboth{IEEE Robotics and Automation Letters. Preprint Version. Accepted 
January, 2019}
{Tinchev \MakeLowercase{\textit{et al.}}: Deep Laser Localization on a CPU} 

\maketitle

\begin{abstract}

  Localization in challenging, natural environments such as forests or woodlands
  is an important capability for many applications from guiding a robot 
  navigating
  along a forest trail to monitoring vegetation growth with handheld sensors. 
  In this work we 
  explore laser-based localization in both urban and natural 
  environments, which is suitable for online applications. We propose a deep 
  learning 
  approach capable of learning meaningful descriptors directly from 3D point 
  clouds by comparing triplets (anchor, positive and negative examples). 
  The 
  approach
  learns a feature space representation for a set of segmented point clouds that 
  are matched between a current and previous observations.
  Our learning method is tailored towards loop closure detection resulting in a 
  small 
  model 
  which can be deployed using 
  only a CPU. The 
  proposed learning method would allow the full pipeline to run on 
  robots with limited computational payload such as drones, quadrupeds or 
  UGVs.
\end{abstract}


\begin{figure}[t!]
	\centering
	\includegraphics[width=.95\columnwidth]{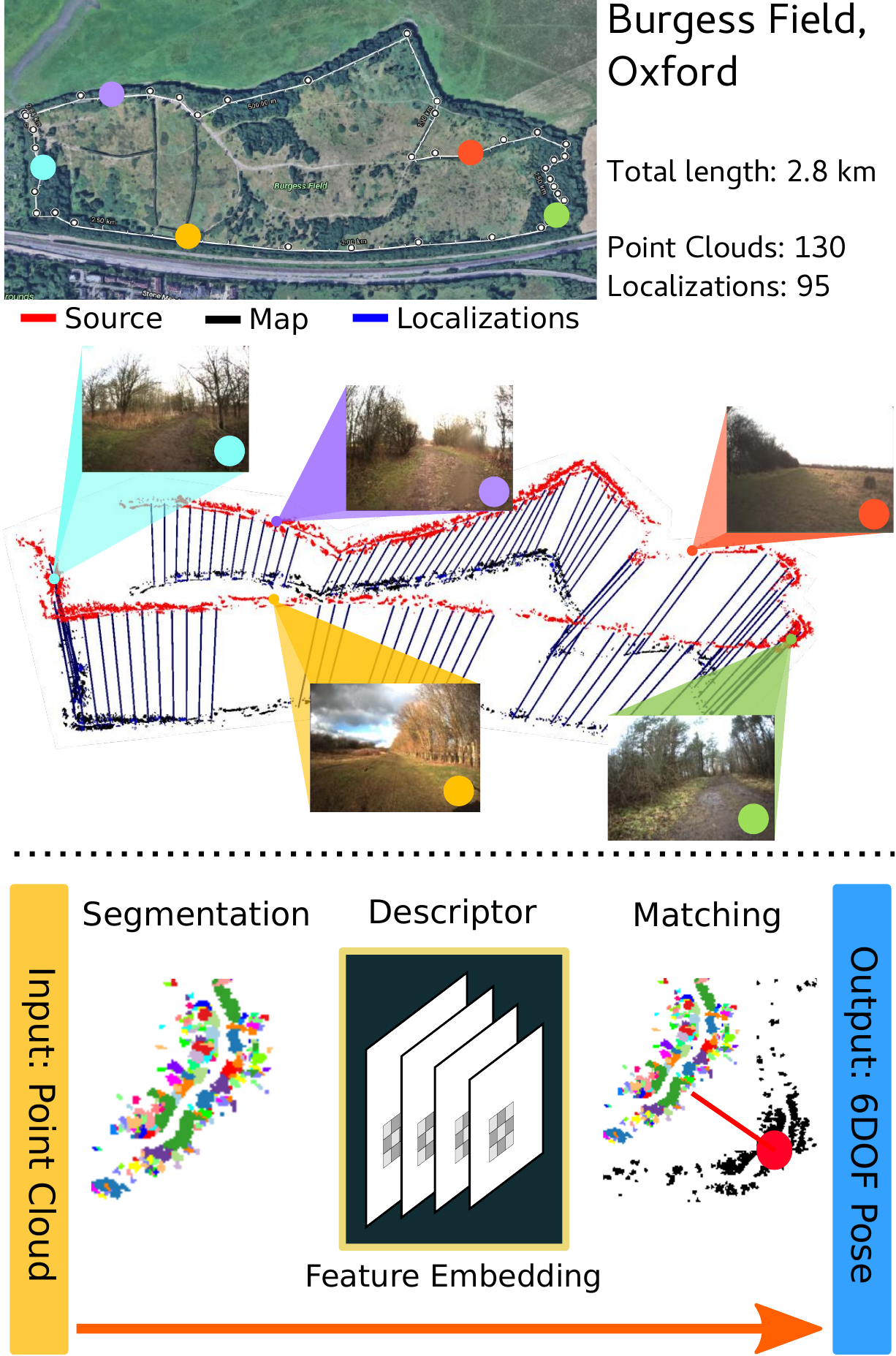}
	\caption{
	\textbf{Top:} Depiction of the obtained results on large, unstructured 
	environments.
	The proposed approach regularly proposes correct loop closures in these 
	challenging natural scenes, using a deep learning architecture running only 
	on a CPU. 
	\textbf{Bottom:} Overview of the proposed methodology. Segments are extracted
	from the input point cloud (coloured section) and the map cloud (in 
	black on top). Our neural 
	network features are extracted for all segments. The features from the map 
	are stored in a database. The features from the query cloud are matched 
	against 
	the database. The final pose is then estimated with PROSAC using both the 
	probability of a segment matches(from the network) and the position of the 
	segments (from the map).
	}
	\label{fig:teaser}
	\vspace{-7mm}
\end{figure}

\section{Introduction}
\label{sec:intro}

\IEEEPARstart{L}{ocalization} is a fundamental task in robotic perception, a 
robot needs
to know where it is to navigate in the environment and to make decisions. It has 
been heavily explored
with computer vision, demonstrating
impressive results at large scales~\cite{mur2015orb, ChurchillIJRR2013, 
AgarwalRome2011}. These 
types of approaches typically assume a certain inherent structure in the 
scene, image features are dependant on repeatable camera 
viewpoint~\cite{AgarwalRome2011}, and methods are often tested in urban 
environments which guide the robot along the route in question~\cite{geiger2013vision}.
While the aforementioned visual approaches have many promising characteristics,
here we explore LIDAR due to its robustness to varying lighting conditions,
changes in viewpoint, and trackline offsets. It is a precise and long range 
sensing modality.

The SegMatch system~\cite{dub2017icra} proposed a modular segment-based 
approach 
for LIDAR teach-and-repeat, which could localize within a prior map
while also retaining a degree of semantic meaning, as the segments matched corresponded to large
physical objects such as cars and parts of a building. The approach was further improved with
learned feature descriptors~\cite{segmap2018} which achieved greater accuracy. The approach 
uses a GPU to achieve real-time performance. 

We are motivated to localize in natural 
scenarios like forests. In these environments
many assumptions about the appearance or the geometry of landmarks are no 
longer valid - vegetation grows between seasons and there is no planar structure. 
Our previous work showed promising results in both structured (urban)
and unstructured (natural) environments~\cite{tinchev18seeing}. While reliable 
localization was achieved, we relied on a hand-crafted set of features which 
limited the performance. 

The main contribution of this work is a novel description method for 
segment-based LIDAR localization. We aim to improve upon previous works by 
learning segment 
representations in a way that inherently handles the variability of the given environment. 
The proposed approach learns a descriptor space which efficiently represents
the similarities between partial observations of the same segment which makes it
robust to incomplete data. We use a neural network to learn this high dimensional 
feature space. The proposed approach utilizes the convolution 
operation proposed in~\cite{li2018pointcnn} to learn an embedding space for both 
urban and natural scenarios directly from the raw point cloud data. The neural 
network has the following characteristics:

\begin{itemize}   
\item Unordered point clouds as input: does not require a
specific ordering of the point clouds in a segment. This makes our approach 
flexible
by avoiding the computational cost of creating specific structures for the
point cloud and aligning the inputs to a grid or voxelization.   
\item Feature space is capable of
being generalized: experimental validation has proven that our proposed deep
learning solution creates a feature space that can generalize,
without the need to be retrained to a new sensor or a different environment.
\item The network can estimate the quality of a match: a probability
is computed and can be used when carrying out probabilistic
geometric validation such as PROSAC~\cite{prosac2005}, making our
approach more efficient.
\end{itemize}

In the results section we demonstrate that our proposed method significantly 
outperforms other hand-engineered approaches, while also improving the 
computational speed in comparison to other deep learning approaches.
In particular we achieve localization performance similar to 
SegMap~\cite{segmap2018}, but do not require a GPU at runtime. Instead the method 
can be deployed online on the CPU of a mobile robot. This performance improvement 
comes in large part because our network is more specifically tailored to the task 
of localization. As presented in \cite{segmap2018},
the SegMap learning approach can be used to compress and reconstruct point clouds
as well as extract and use semantic meaning from the segments to aid 
localization. We demonstrate performance in urban environments as well as natural 
environments so as to demonstrate our approach's generality and robustness.

\section{Related Work}
\label{sec:related}


Robot localization has been heavily explored using different sensors
such as LIDAR~\cite{Thrun99montecarlo}, vision~\cite{furgale_jfr10},
GPS~\cite{rss_Levinson07} or radar~\cite{radar01tro} with reliable approaches
often combining different sensing modalities~\cite{Thrun99montecarlo, 
rss_Levinson07, radar01tro}. 
Our proposed work focuses on global LIDAR localization particularly applied to 
unstructured environments
rather than incremental localization or odometry. Here we will briefly overview 
some methods performing LIDAR localization. 

For self-driving
vehicles many approaches have taken advantage of LIDAR reflectivity to achieve
precise localization. These methods are commonly used in commercial approaches,
for example, the approach of~\cite{wolcottIjrr17} uses a prior map of reflectivity
and exploits road marks to
reliably localize a vehicle in an urban environment. The authors formulate the
world as a mixture of Gaussians (GMM) over 2D grid cells. The GMM represents the 
heights of points in each cell and the reflectance in a vigorous way that allows 
the approach to be robust to weather alteration and road degradation.

\begin{figure*}[t]
	\vspace{0.3cm}
	\centering
 	\includegraphics[width=\textwidth]{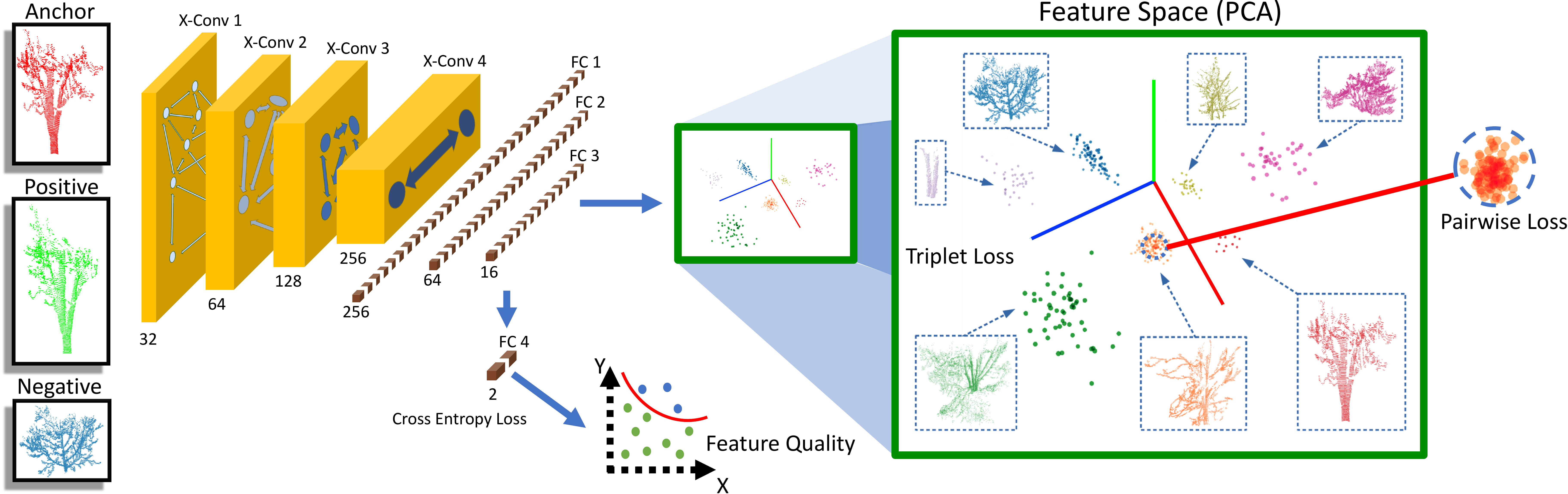}

	\caption{ Network architecture used for learning the feature embeddings (left) and a
	visualization of the embedding space after training (right). The network takes 
	as input raw point data, it uses four x-convolutional layers 
	($\mathcal{X}$-conv) and three fully 
	connected (FC) layers to estimate the feature space. A feature quality branch is also created 
	by appending a fully connected layer to the feature embedding. The 
	networks learns to 
	cluster similar objects (denoted in the same colour) close together while separating 
	dissimilar objects. The final feature descriptor is trained using a combination of triplet and 
	pairwise losses.
	}
	\label{fig:network_architecture}
\vspace{-5mm}
\end{figure*}

When localization is carried out concurrently with mapping it is often treated as 
a research
field of its own (Simultaneous Localization and Mapping, SLAM) where the key 
problem of loop-closure detection is often equivalent to global localization.
In this context the robot's odometry
provides a rough estimate of the position of the robot within the world which
can be leveraged to avoid searching for loop closures over the entire map, thus 
keeping the
computational cost low~\cite{hessICRA16}. When the assumption of a local 
neighbourhood is lost
most of these approaches require ICP alignments or similar costly 
alternatives~\cite{pomerleau13ar}. 

Recent approaches aim to localize on a higher
level than on explicit point-to-point basis. The concept of segment localization 
was presented in~\cite{original_segment_paper2012} but more specifically our work 
is directly motivated
by the works of Dub\'{e}~\textit{et al}~\cite{dub2017icra, segmap2018}.
We will describe the framework of segment-based localization in 
Section~\ref{sec:approach}. Features created from LIDAR data are less informative 
or unique than visual features,
thus approaches often prune the candidate matches to reduce the
percentage of outliers before performing a robust geometric
validation~\cite{zachCVPR15}. One of the key advantages of using segments
instead of keypoints is that more semantically meaningful entities are
extracted, increasing the repeatability of the descriptors. This was
demonstrated in~\cite{dub2017icra} and was applied to urban environments. 
In~\cite{segmap2018} segment localization evolved into a segment-based SLAM
that introduced a new segment feature that also encoded semantics and
volumetric shape. This interesting line of work learns features suitable for 
localization of voxelized segments and the semantics of the map simultaneously.
The approach was evaluated in realistic urban setting and in derelict buildings.
Elbaz~\textit{et al}~\cite{elbaz3d} used segments produced from a Random Sphere 
Cover Set - overlapping point cloud spheres where each point of the original 
point cloud 
could be part of more than one sphere. These spheres were then projected to 2D 
depth images and processed by a deep auto-encoder. In our previous 
work~\cite{tinchev18seeing} 
we proposed a larger hybrid feature
descriptor which improved detection performance and achieved localizations in
unstructured environments such as forests, while also being capable of handling 
the urban scenarios.


The main drawback of classical approaches that use LIDAR data is that
they rely on handcrafted features which do not perform as well as learned
features~\cite{alexNet}. When trying to apply learning techniques to 3D data
there are two prominent approaches - representing the data as a voxelized grid, 
or 
by using the more common representation of point clouds. Initially
these approaches tended to use voxelized inputs as the tensor operators
generalize directly to this representation and achieve promising results as
shown in~\cite{choy20163d}. The main drawback of this approach is the amount of 
computation required to model the entirety of the environment.
In~\cite{choy20163d} the authors used a 32x32x32 voxelized grid, resulting in a 
dimensionality of 32768, with 3D convolutions applied in
strides. Because the input resolution is low (due to the
computational complexity), these approaches 
are not capable of handling detailed shapes such as the legs of a chair, for 
example, as shown in~\cite{haoqiang2017pointsetNet}.
The issue of handling point clouds was addressed in~\cite{qi2016pointnet,
qi2017pointnetplusplus} by learning a symmetry function approximated by a
multi-layer perceptron (MLP). The authors created a neural network layer that was
invariant to point order. In contrast to the aforementioned volumetric approach, 
a raw processed point cloud containing 1024 points resulted in a dimensionality 
of only 3072. In addition, by using a MLP as the basis of their
layer and by aggregation of spatial data, Li~\textit{et al}~\cite{li2018pointcnn} 
managed to create a more descriptive layer to handle unordered point clouds. 

In~\cite{uypointnetvlad} a query point cloud is downsampled and matched against a 
collection of previously collected submaps globally. All clouds are described on 
a GPU using a combination of networks trained with a lazy quadruplet 
loss to produce a 256-dimensional feature vector. In contrast, we firstly segment 
the query and map point clouds to produce a small set of segments, local 
keypoints. The segments are then described in real time on a CPU. Thus, our 
approach uses a combination 
of local features for each segment and matches those in the map. In this way we 
are robust to disturbances from occlusions and changing environments.

\section{Methodology}
\label{sec:approach}

In this section we present Efficient Segment Matching (\textbf{ESM}). This work 
retains the approach to pose estimation which was initially proposed 
by~\cite{dub2017icra}. Our contributions in this paper are 1) learning a novel 
segment descriptor, 2) together with a learned per 
descriptor measure of performance which 
enables the pruning of the features before matching and 3) the 
use of a probabilistic robust pose estimation~\cite{prosac2005} to improve 
matching performance.

The problem 
is formulated as follows: given an input point cloud $\mathcal{P} 
=\{p_i \in \mathbf{R}^3\}$ 
and a prior map $\mathcal{M} 
=\{p_j \in \mathbf{R}^3\}$, we wish to estimate the pose of 
$\mathcal{P}$ in the map $\mathcal{M}$. We split the task into 
four modules - Segmentation, 
Description, Matching and Pose Estimation. Our methodology is illustrated 
in~\figref{fig:teaser}, bottom. An input point 
cloud 
is first segmented using Euclidean segmentation to produce a 
number of 3D point cloud objects based on the distance of points between them. 

Next the segments are passed to a neural network to
extract a descriptor of each segment and a measure of quality of that 
descriptor. For each segment in the live point cloud the $N$-closest neighboring 
segments in the map
are found based on the Euclidean distance of the descriptors. These matches are 
ordered based on the extracted 
feature
quality and passed to PROSAC~\cite{prosac2005} to produce a set of
possible localizations. In this work the ordering is done by computing the joint
probability of both features (the feature from the live point cloud and the one 
from the map). The order of the matches is important because the 
geometric consistency and PROSAC consider first the matches with higher likelihood.
Each module of this pipeline is described in depth in previous 
works~\cite{tinchev18seeing,dub2017icra}. In the following section we 
detail the architecture and the learning approach.

\subsection{Architecture}

Our network architecture is constructed based on the novel convolutional layers 
presented in~\cite{li2018pointcnn}.~\figref{fig:network_architecture} illustrates 
our model.
The network learns directly from point cloud data, it takes as input a batch of raw
point cloud segments. Each segment is uniformly downsampled to 256 points, 
zero-centered with normalized variance. 

The network used for our descriptor 
learning approach 
consists of four $\mathcal{X}$-conv
operators~\cite{li2018pointcnn} and three fully connected layers; 
dropout~\cite{dropout2014} of 0.5 
is applied at the second fully connected layer. The outputs ot the last fully 
connected layer are used as the descriptor.
The $\mathcal{X}$-conv operator convolves local regions, similar to
convolutions in images by a CNN. For each point, its closest $N$ neighbors are
projected to a local coordinate frame and lifted to a higher dimensional space
with a multi-layer perceptron (MLP). The $\mathcal{X}$-conv operator learns a
convolution based on the MLPs of the neighbouring points. In order for the top
$\mathcal{X}$-conv operators to see a larger portion of the point cloud a
dilation rate ($D$) is applied~\cite{dilatedConv2016}. In this way the receptive 
field of the top layers is increased without an increase of the neighbouring 
point size ($N$) or the kernel size.

We have selected a specific configuration of the dilation and neighbourhood
in each of our layers that better represents the problem we are trying to
solve. The first layer uses a neighbourhood of 8 and a dilation of 1 
per $\mathcal{X}$-conv operator, the second a neighbourhood of 12 and dilation 3,
the third a neighbourhood of 16 and a dilation of 3 and the fourth a neighbourhood
of 16 and a dilation of 4. The reason for this is that the first layer will look
at a few points (neighbourhood 8) that are immediately close (dilation 1). The 
second 
will look into more points (neighbourhood 12) further out (dilation 3). This
way the network creates a representation that aggregates the information slowly
creating a hierarchical representation of the whole segment. This is done for each
point so as to create a feature of the entire segment from the relative viewpoint
of each point. The final descriptor is computed as an average of the features for all the
points in a single point cloud. In this way, the feature fuses data from 
different points creating an expressive yet simple representation.
In our experiments we managed to achieve good performance by using an 
embedding size 
of 16 dimensions. The dimensionality of a descriptor has a strong impact on
the performance of matching. This keeps the computational cost low even if the 
segmentation algorithm produces many more
segments.

The proposed network architecture also includes a classification branch that estimates 
a measure of the quality of each descriptor. To be able to train this branch we 
need to train the feature
network first. When training the classification branch the descriptor 
layers are not modified, the model optimizes a single fully connected layer of 
size two. This represents a logistic regression that classifies whether a feature 
is good for 
matching or 
not. The quality of the feature is determined during training if a 
successful 
match is found within the first $K$ neighbors in the dictionary. The 
classification branch 
is trained until its accuracy converges at $\approx 70\%$ with $K=1$ neighbor. 
At test time the confidence score from the classification branch is used to 
compute 
a joint probability of all matched segments. Since two matched segments are 
gathered 
during independent observations, we simply multiply the 
probabilities. During the last stage of the 
localization 
pipeline the candidates are sorted using the joint probability distribution and the pose 
is estimated using PROSAC~\cite{prosac2005} with applied geometric consistency 
constraints~\cite{tinchev18seeing}.

\subsection{Learning the Segment Feature Space}

To learn the feature space in the proposed architecture a variation of the triplet 
loss~\cite{tripletloss_NIPS2005} is used. The triplet loss clusters together 
samples in the 
feature 
space  that have been labelled as similar and tries to separate samples 
that have been
labelled as different. By applying the same label to different examples of a
segment we introduce invariance to many factors such as noise in the measurements 
or 
incomplete segments. The triplet loss defines a triplet as a combination of a sample 
(anchor) with other two samples, one with the same label and the other with a different 
label (\figref{fig:network_architecture}, left). A pairwise term is defined as a 
combination
of a sample (anchor) with a different sample that has the same label. To train our descriptor
we use a variation of the triplet loss similar to the one defined 
in~\cite{wohlhart2015learning} - 
in our case we use a squared $L_2$ norm as shown in~\eqref{eq:combined_loss} 
while they use a 
regular 
$L_2$ norm. To train the model we use a batch size of $256$ point clouds from 
which 
we extract a
large set of triplet and pairwise terms. 
\begin{eqnarray}
\label{eq:combined_loss}
\mathcal{L} = \mathcal{L}_{triplets} + \alpha\mathcal{L}_{pairs} + \lambda\sqnorm{w}.
\end{eqnarray}
During training we identify $\approx300$ times more triplet than pairwise
terms. We balance this effect 
using the $\alpha$ parameter. In~\eqref{eq:combined_loss} $w$
denotes the parameters of our model, weighted by $\lambda=10^{-6}$. The proposed 
architecture 
contains only $300$\,K trainable parameters. We use an
initial learning rate of $\eta=10^{-3}$, decaying to minimum $\eta=10^{-6}$ with
ADAM~\cite{kingma2014adam} as optimizer. We describe the two losses in detail below.

The classification network is trained using the same data used for the embeddings
layer. The features are precomputed and used as a dictionary. To train the
classification branch, for each sample we find the $K$ closest matches in the
dictionary. A binary label is assigned for each sample: if a match is found
in the dictionary it is labelled as 1, otherwise 0. The network optimizes a
softmax cross-entropy loss for the classification branch, given the aforementioned
labels.

\subsection{Triplet Loss}

As commented before, we use a variation of the definition of triplet loss similar to the one
in~\cite{wohlhart2015learning}. This loss modified the original triplet loss to 
solve the vanishing gradient problem. We modify this loss by using squared $L_2$ 
norm, instead of the standard $L_2$ norm,
as it created a better clustering of the feature space. The final 
triplet loss is defined as the sum of the following cost function over all the 
triplets:
\begin{eqnarray}
\label{eq:triplet_loss}
\mathcal{L}_{triplets} = \sum_{(s_i, s_j, s_k)\in \mathcal{T}}{c(s_i,s_j,s_k)}
\end{eqnarray}
\begin{eqnarray}
\label{eq:triplet_cost}
c(s_i,s_j,s_k) = \max \Big(0,1-\frac{\sqnorm{f_w(x_i) - f_w(x_k)}}{\sqnorm{f_w(x_i)-f_w(x_j)}+m}\Big)
\end{eqnarray}
$\mathcal{T}$ denotes the set of all possible
triplets. $s_i$ and $s_j$ are segments with the same label, while $s_i$ and $s_k$ are
dissimilar. $f_w(x_{i})$ is the output of the last
descriptor layer for an input point cloud $x_{i}$ and $m$ is a margin 
regularizer. The latter
denotes the minimum ratio for the Euclidean distances between dissimilar pairs
of point clouds and similar ones. We use $m=0.01$ in our experiments. In this
manner the triplet loss will cluster similar objects together, while separating
dissimilar ones to be farther apart in the feature
space. The proposed architecture will provide a descriptor of the whole segment per point, centered at each
of those points. Our feature is computed as the average of those descriptors. Our 
losses are 
computed using the averaged features.~\figref{fig:network_architecture} (right) 
presents a PCA visualization
of the feature space after the network has been trained with the corresponding
segments. Different instances of the same tree are clustered together while
other trees form their own clusters.

\begin{figure}[t!]
	\centering
	\begin{subfigure}[b]{0.48\textwidth}
		\includegraphics[width=\textwidth]{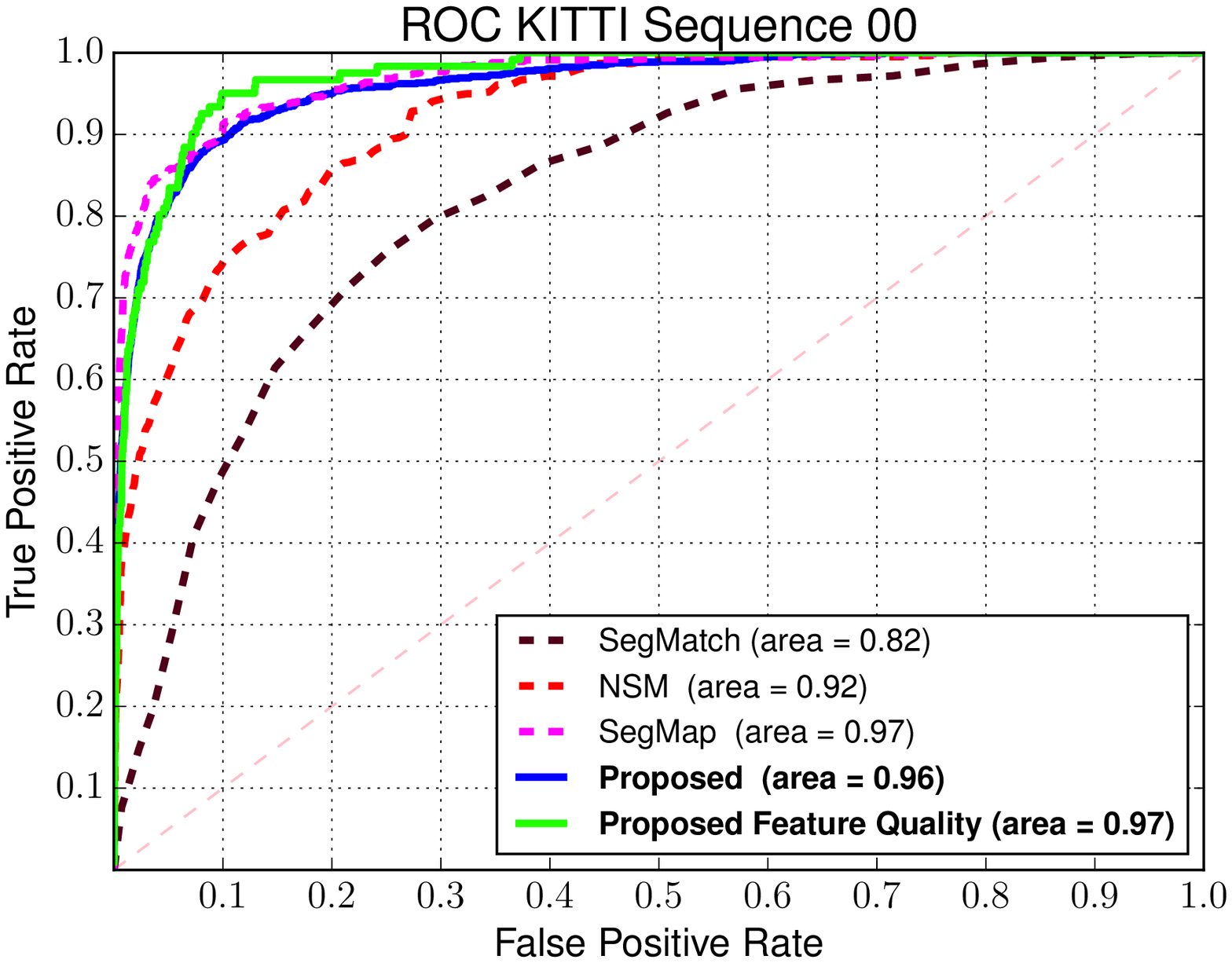}
	\end{subfigure}
	
	\begin{subfigure}[b]{0.48\textwidth}
		\includegraphics[width=\textwidth]{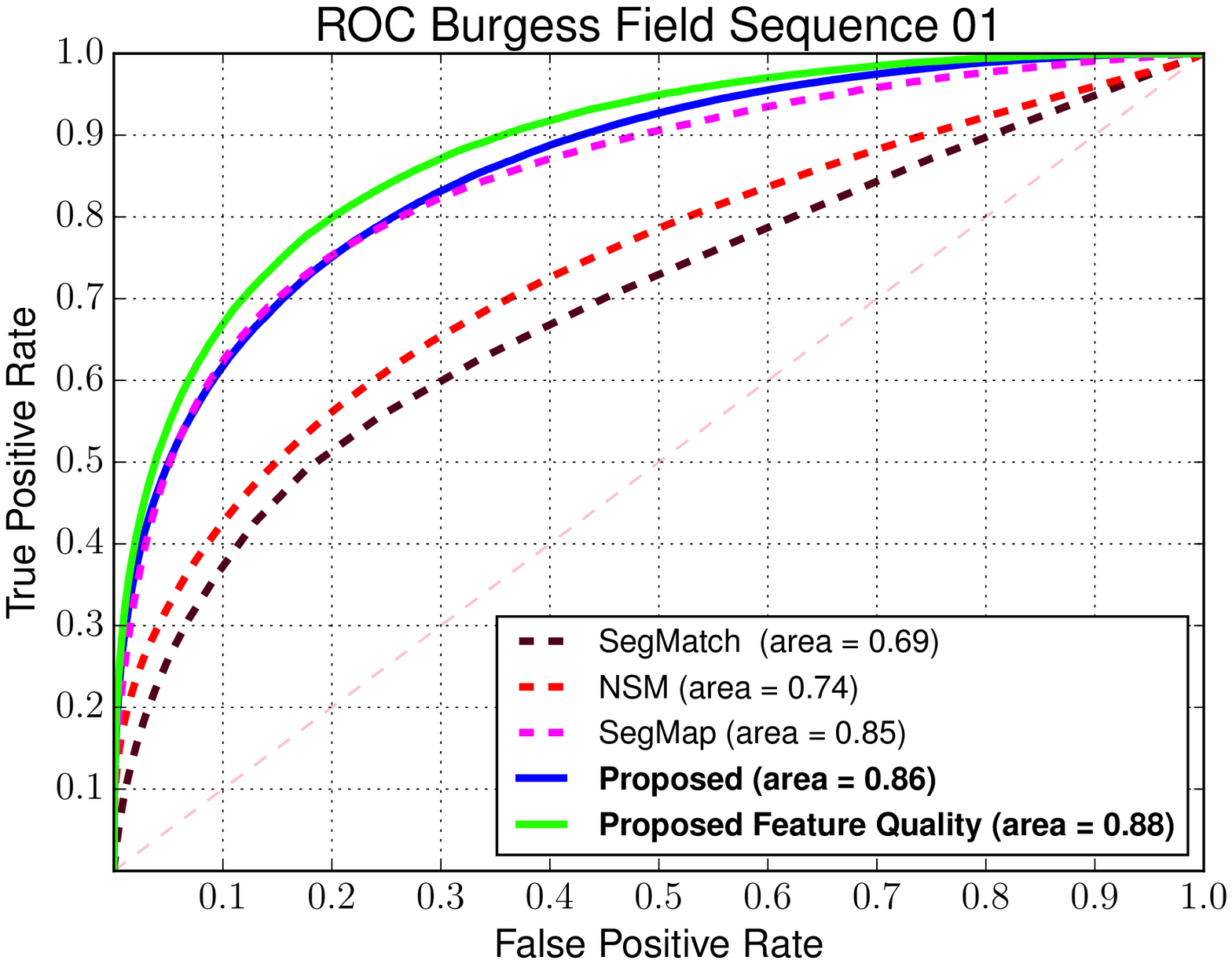}
	\end{subfigure}
	
	\caption{Receiver Operating Characteristic (ROC) curves for the 
		proposed and baseline approaches~\cite{dub2017icra, segmap2018, 
			tinchev18seeing} on urban (top) and natural (bottom) datasets. The 
			proposed 
		approach performs comparably to the current 
		state of the art on both environments. Learning approaches 
		outperform hand-engineered ones in both urban and natural scenarios.}
	\label{fig:roc}
	\vspace{-6mm}
\end{figure}

\subsection{Pairwise Loss}

The pairwise loss minimizes the distance between samples of the same class
(\figref{fig:network_architecture}, orange). As proposed by~\cite{wohlhart2015learning},
we optimise the following:
\begin{eqnarray}
\label{eq:pairwise_loss}
\mathcal{L}_{pairs} = \sum_{(s_i, s_j)\in \mathcal{P}}{\sqnorm{f_w(x_i) - f_w(x_j)}}
\end{eqnarray}
where $\mathcal{P}$ denotes the set of all pairs. The
loss aids the training process by generating very tight clusters, which in
turn improve the KNN retrieval during the matching phase.

\section{Experimental Evaluation}
\label{sec:exp}

The main focus of this work is to utilize learned features in point clouds to 
localize with respect to a prior map. The experiments are designed to support our 
key claims:

\begin{itemize}
	\item A novel learned feature descriptor that generalizes across a variety of 
	natural and urban datasets.
	\item Real time operation on a CPU as a result of a compact network 
	architecture.
	\item The approach produces a measure of quality for each feature which can 
	be used during the pose estimation step to decrease computation.
\end{itemize}

We provide comparisons against a popular method for segment-based localization, 
SegMatch\footnote{\label{fn:impl}We use the open-source implementations of 
SegMatch/SegMap.}, as 
proposed in~\cite{dub2017icra}, a data-driven 
incremental approach 
SegMap\footnoteref{fn:impl}~\cite{segmap2018}, and our previous work 
NSM~\cite{tinchev18seeing}.  As a 
metric to evaluate the approaches we have chosen to compare the True Positive 
Rate (TPR) against the False Positive Rate (FPR) for each classifier, the number 
of localizations on each dataset, and the computation time for each pipeline.
Supplementary material about our model's hyper-parameters and a video accompany
the paper at~\url{http://ori.ox.ac.uk/esm-localization}.

\begin{figure*}[t]
	\vspace{0.2cm}
	\centering
	\includegraphics[width=0.92\textwidth]{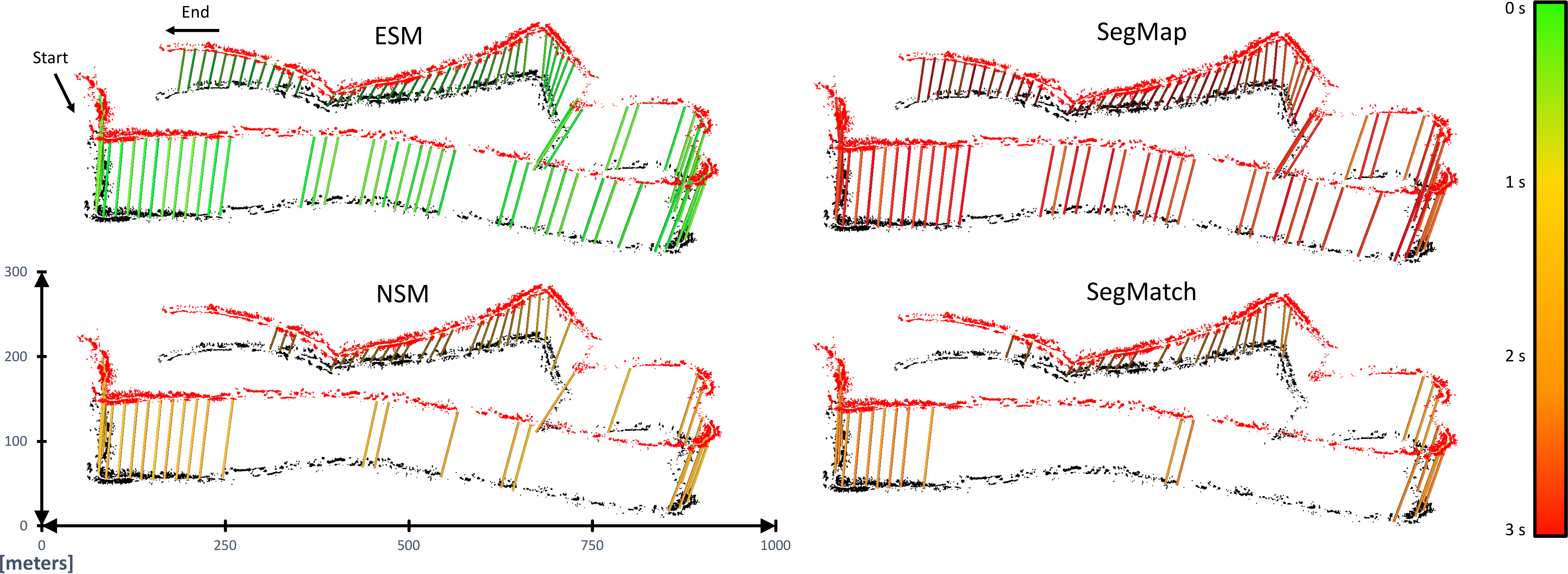}
	\caption{Illustration of the localization performance of our approach (top 
		left) compared 
		to the baselines on the Burgess Field dataset. The current point clouds 
		(red) are registered to a 
		previous 
		map (black). Each vertical line corresponds to a recognised place, while 
		its 
		color corresponds to the time taken to perform the localization.}
	\label{fig:bf_localizations}
\vspace{-3mm}
\end{figure*}

\begin{table*}[t]
	\centering
	\begin{tabular}{|l|c|c|c|c|c|c|c|c|c|}
		\hline
		\cline{1-5} \cline{7-10}
		\multicolumn{5}{|c|}{\cellcolor[HTML]{FFCC67}{\color[HTML]{000000} 
				\textbf{Dataset Characteristics}}}  & 
		\multicolumn{1}{l|}{\cellcolor[HTML]{EFEFEF}\textbf{}} & 
		\multicolumn{4}{c|}{\cellcolor[HTML]{FFCC67}{\color[HTML]{000000} 
				\textbf{Number of 
					Localizations}}}                                              
					            
		\\
		\hline
		\rowcolor[HTML]{EFEFEF} 
		\multicolumn{1}{|c|}{\cellcolor[HTML]{EFEFEF}\textbf{Name}} & 
		\textbf{Type} & \textbf{Length (meters)} & \textbf{Sensor}        & 
		\textbf{Num. Clouds} & \cellcolor[HTML]{EFEFEF}\textbf{} & 
		\textbf{SegMatch} & \textbf{NSM} & \textbf{SegMap}                     & 
		\textbf{ESM}                       \\ \hline
		\cellcolor[HTML]{EFEFEF}\textbf{KITTI}                      & 
		Urban         & 1300                     & Accum. Velodyne-HDL64E & 
		134                  & \cellcolor[HTML]{EFEFEF}          & 
		41                & 53           & 62                                  & 
		\cellcolor[HTML]{9AFF99}\textbf{63}  \\ \hline
		\cellcolor[HTML]{EFEFEF}\textbf{George Square}              & City 
		Park     & 500                      & Push-broom             & 
		30                   & \cellcolor[HTML]{EFEFEF}          & 
		8                 & 9            & 8                                   & 
		\cellcolor[HTML]{9AFF99}\textbf{11}  \\ \hline
		\cellcolor[HTML]{EFEFEF}\textbf{Cornbury Park}              & 
		Forest        & 700                      & Velodyne-HDL32E        & 
		1125                 & \cellcolor[HTML]{EFEFEF}          & 
		N/A               & 29           & 231                                 & 
		\cellcolor[HTML]{9AFF99}\textbf{248} \\ \hline
		\cellcolor[HTML]{EFEFEF}\textbf{Burgess Field}              & 
		Forest        & 2800                     & Push-broom             & 
		130                  & \cellcolor[HTML]{EFEFEF}          & 
		41                & 54           & \cellcolor[HTML]{9AFF99}\textbf{95} & 
		94                                   \\ \hline
	\end{tabular}
	\caption{Number of localizations across one urban and three natural 
		datasets. The proposed approach performs comparably to the state of the 
		art. These experiments also show the ability of the learning approaches 
		to handle data from different types of LIDAR sensors.}
	\label{tab:loc_results}
\vspace{-5mm}
\end{table*}

\subsection{Datasets}

We perform evaluations using three datasets of our own which were collected in 
natural environments as well as on a publicly available dataset. First, we 
compared our novel descriptor against the baseline approaches using the 
KITTI~\cite{geiger2013vision} dataset (Karlsruhe, Germany), taken by a 3D 
Velodyne-HDL64 sensor in an urban scenario. We use Sequence 06 and 05 to 
train all descriptors, and Sequence 00 to test them. We also compare the 
localization performance on Sequence 00. Second, we compared localization 
performance in a park and forest datasets - George's Square (Edinburgh, 
Scotland) and Cornbury Park (Oxfordshire, UK). The former was captured by 
a Clearpath Husky UGV equipped with a SICK LMS511 LIDAR, while the latter 
was captured by a vehicle equipped with a Velodyne HDL32E. The datasets are 
described in detail in~\cite{tinchev18seeing}. 

Finally, we used a dataset collected in a natural environment, located at Burgess 
Field (Oxfordshire, UK). The dataset was captured 
by a Clearpath Husky UGV equipped with a Bumblebee3 forward facing camera, a 
Velodyne VLP-16 LIDAR and a Push-broom LIDAR LMS151. The vehicle traversed the 
same loop of $\approx 2.8$\,km twice a month during a span of six months. We used 
the 
Velodyne data of Sequence 02 and Sequence 01 for training and testing the 
classifiers. These two sequences were collected two weeks apart each other in 
February when the foliage was shed. We built a prior 
map from the Push-broom LIDAR of Sequence 02 
using~\cite{ChurchillIJRR2013}. The source swathes were built using the VO 
method~\cite{ChurchillIJRR2013} 
for every 22 meters the vehicle traversed in Sequence 01.

To label the data we combined four consecutive Velodyne scans using the 
motion estimator from~\cite{ChurchillIJRR2013} and extracted segments using 
distance-based segmentation algorithm. All segments to be within $0.5$\,m in consecutive 
point clouds are considered the same object and labelled as such. All segments 
further than $4.0$\,m apart are considered a non-match. In this manner we extracted a 
total of $82871$ segments across $38714$ classes for training and $45409$ segments in 
$20583$ classes for testing. Training our model on Burgess Field takes 3 hours and 25 
minutes from scratch. We have used this model to evaluate our approach on all 
natural datasets. For experiments carried out on urban datasets, we trained a different model on 
KITTI in 11 hours and 45 minutes. For all the experiments and methods we use 
Euclidean segmentation. We segmented the data into individual point clouds depending 
on the sampling density: 3D Velodyne data: $200$--$15000$ points due to higher 
frequency, 
Push-broom type LIDAR: $200$--$50000$ points due to higher density.

\subsection{Classifier Performance}

The first experiment evaluates the performance of our model, with and without the 
feature quality, on KITTI and Burgess Field datasets.~\figref{fig:roc} illustrates the Receiver 
Operator Characteristic (ROC) curves for each of the algorithms. The ROC curve 
for the feature quality network is created by removing the top \textit{bad} 
matches prior to evaluation. A performance decrease is 
seen between the two datasets for all algorithms. We attribute this to the more 
challenging natural structure in the second environment. In brief, our learning 
approach generalizes better than engineered features (NSM, SegMatch)
across datasets and performs comparably to the learned approach (SegMap). Pruning 
the matches based on the feature quality shows an 
improvement with respect to the basic features. For all models we have set the 
classifier 
thresholds at FPR=$0.1$ for urban and FPR=$0.2$ for forest environments. We did 
not retrain the models for each forested scenario or different sensor modality. 
This shows the proposed features can generalize 
between datasets and sensors.

\begin{table*}[t]
	\vspace{0.3cm}
	\centering
	\begin{tabular}{|
			>{\columncolor[HTML]{EFEFEF}}l |c|
			>{\columncolor[HTML]{EFEFEF}}l |
			>{\columncolor[HTML]{9BFF9B}}c |c|c|c|c|
			>{\columncolor[HTML]{BAFF9B}}c |c|}
		\hline
		\multicolumn{10}{|c|}{\cellcolor[HTML]{FFCC67}\textbf{CPU Multi-core 
				execution 
				times}}

		\\
		\hline
		\textbf{Algorithm    }                         & 
		\multicolumn{1}{c|}{\cellcolor[HTML]{EFEFEF}\textbf{Desc. size}} & 
		\textbf{} & 
		\multicolumn{1}{c|}{\cellcolor[HTML]{EFEFEF}\textbf{Segmentation}} & 
		\multicolumn{1}{c|}{\cellcolor[HTML]{EFEFEF}\textbf{Preprocessing}} & 
		\multicolumn{1}{c|}{\cellcolor[HTML]{EFEFEF}\textbf{Descriptor}} & 
		\multicolumn{1}{c|}{\cellcolor[HTML]{EFEFEF}\textbf{Matching (K)}} & 
		\multicolumn{1}{c|}{\cellcolor[HTML]{EFEFEF}\textbf{Prunning (RF)}} & 
		\multicolumn{1}{c|}{\cellcolor[HTML]{EFEFEF}\textbf{Pose Est.}} & 
		\multicolumn{1}{c|}{\cellcolor[HTML]{EFEFEF}\textbf{Total}} \\ \hline
		SegMatch                                   & 
		7 (647)                                                             &   
		& 
		17                                                                 & 
		0                                                                   & 
		\cellcolor[HTML]{FFF69B}605                                      & 
		\cellcolor[HTML]{BAFF9B}50 (200)                                    & 
		\cellcolor[HTML]{FFF69B}755                                         & 
		64                                                                    & 
		1491\,ms                                                     \\ \hline
		NSM                                        & 
		66                                                                    &   
		& 
		17                                                                 & 
		0                                                                   & 
		\cellcolor[HTML]{9BFF9B}24                                      & 
		\cellcolor[HTML]{E7FF9B}207 (200)                                    & 
		\cellcolor[HTML]{FFF69B}913                                         & 
		65                                                                    & 
		1226\,ms                                                     \\ \hline
		SegMap                                     & 
		64                                                                    &   
		& 
		17                                                                 & 
		\cellcolor[HTML]{9BFF9B}15                                          & 
		\cellcolor[HTML]{FFBE9B}5902                                     & 
		\cellcolor[HTML]{9BFF9B}19 (25)                                    & 
		0                                                                   & 
		\cellcolor[HTML]{9BFF9B}21                                                
		                    & 
		5974\,ms                                                     \\ \hline
		ESM                                      & 
		16                                                                    &   
		& 
		17                                                                 & 
		\cellcolor[HTML]{9BFF9B}3                                           & 
		\cellcolor[HTML]{FFF69B}578                                      & 
		\cellcolor[HTML]{9BFF9B}8 (25)                                     & 
		0                                                                   & 
		\cellcolor[HTML]{9BFF9B}9                                                
		                    & 
		615\,ms                                                      \\ \hline
		\multicolumn{10}{|c|}{\cellcolor[HTML]{FFCC67}\textbf{CPU Single-core 
				execution 
				times}}

		\\
		\hline
		\textbf{Algorithm}                         & 
		\multicolumn{1}{c|}{\cellcolor[HTML]{EFEFEF}\textbf{Desc. size}} & 
		\textbf{} & 
		\multicolumn{1}{c|}{\cellcolor[HTML]{EFEFEF}\textbf{Segmentation}} & 
		\multicolumn{1}{c|}{\cellcolor[HTML]{EFEFEF}\textbf{Preprocessing}} & 
		\multicolumn{1}{c|}{\cellcolor[HTML]{EFEFEF}\textbf{Descriptor}} & 
		\multicolumn{1}{c|}{\cellcolor[HTML]{EFEFEF}\textbf{Matching (K)}} & 
		\multicolumn{1}{c|}{\cellcolor[HTML]{EFEFEF}\textbf{Prunning (RF)}} & 
		\multicolumn{1}{c|}{\cellcolor[HTML]{EFEFEF}\textbf{Pose Est.}} & 
		\multicolumn{1}{c|}{\cellcolor[HTML]{EFEFEF}\textbf{Total}} \\ \hline
		SegMatch                                   & 
		7 (647)                                                             &   
		& 
		21                                                                 & 
		0                                                                   & 
		\cellcolor[HTML]{FFF69B}596                                      & 
		\cellcolor[HTML]{BAFF9B}70 (200)                                     & 
		\cellcolor[HTML]{FFF69B}773                                         & 
		64                                                                    & 
		1524\,ms                                                     \\ \hline
		NSM                                        & 
		66                                                                    &   
		& 
		21                                                                 & 
		0                                                                   & 
		\cellcolor[HTML]{9BFF9B}37                                       & 
		\cellcolor[HTML]{E7FF9B}213 (200)                                   & 
		\cellcolor[HTML]{FFF69B}936                                         & 
		65                                                                    & 
		1272\,ms                                                     \\ \hline
		SegMap                                     & 
		64                                                                    &   
		& 
		21                                                                 & 
		\cellcolor[HTML]{9BFF9B}26                                          & 
		\cellcolor[HTML]{FF9B9B}25945                                    & 
		\cellcolor[HTML]{9BFF9B}26 (25)                                    & 
		0                                                                   & 
		\cellcolor[HTML]{9BFF9B}21                                                
		              
		      & 
		26039\,ms                                                    \\ 
		\hline
		ESM                                      & 
		16                                                                    &   
		& 
		21                                                                 & 
		\cellcolor[HTML]{9BFF9B}3                                           & 
		\cellcolor[HTML]{FFE19B}2126                                     & 
		\cellcolor[HTML]{9BFF9B}12 (25)                                     & 
		0                                                                   & 
		\cellcolor[HTML]{9BFF9B}11                                                
		                    & 
		2173\,ms                                                     \\ \hline

		\multicolumn{10}{|c|}{\cellcolor[HTML]{FFCC67}\textbf{GPU Computation 
		comparison}}

		                                                \\
		 \hline
		\rowcolor[HTML]{EFEFEF} 
		\textbf{Algorithm}                        & 
		\textbf{Desc. size}                                                       
		     & 
		\multicolumn{1}{c|}{\cellcolor[HTML]{EFEFEF}} & 
		\multicolumn{1}{c|}{\cellcolor[HTML]{EFEFEF}\textbf{Segmentation}}

		         &
		 \textbf{Preprocessing}                                                   
		  & 
		\textbf{Descriptor}                                                     & 
		\textbf{Matching}                                                         
		   & \textbf{Pruning (RF)} &
		\multicolumn{1}{c|}{\cellcolor[HTML]{EFEFEF}\textbf{Pose 
		Est.}}                                      &
		{\textbf{Total}}

		\\
		 \hline
		\cellcolor[HTML]{EFEFEF}SegMap   & 
		64                                                                    & 
		\multicolumn{1}{c|}{\cellcolor[HTML]{EFEFEF}}                         & 
		\multicolumn{1}{c|}{\cellcolor[HTML]{9BFF9B}17}            & 
		{\cellcolor[HTML]{9BFF9B}15}                                      
		                     
		               &
		 \cellcolor[HTML]{9BFF9B}15                                       & 
		\cellcolor[HTML]{9BFF9B}19                                     & 
		0                                         & 
		\multicolumn{1}{c|}{\cellcolor[HTML]{9BFF9B}15}     &
		{81\,ms}

		                                                          \\ \hline
		\cellcolor[HTML]{EFEFEF}ESM     & 
		16                                                                    & 
		\multicolumn{1}{c|}{\cellcolor[HTML]{EFEFEF}}                         & 
		\multicolumn{1}{c|}{\cellcolor[HTML]{9BFF9B}17} 
		&            
		\multicolumn{1}{c|}{\cellcolor[HTML]{9BFF9B}3}                            
		                                
		               &
		 \cellcolor[HTML]{9BFF9B}2                                        & 
		\cellcolor[HTML]{9BFF9B}8                                     & 
		0                                           & 
		\multicolumn{1}{c|}{\cellcolor[HTML]{9BFF9B}6}              &
		36\,ms

		                                                          \\ \hline
		
	\end{tabular}
	\caption{Average computational times in milliseconds recorded per point cloud 
		on the Burgess Field dataset.}
	\label{tab:performance}
\end{table*}

\subsection{Localization Performance}

\begin{figure*}[t]
	\centering
	\includegraphics[width=0.92\textwidth]{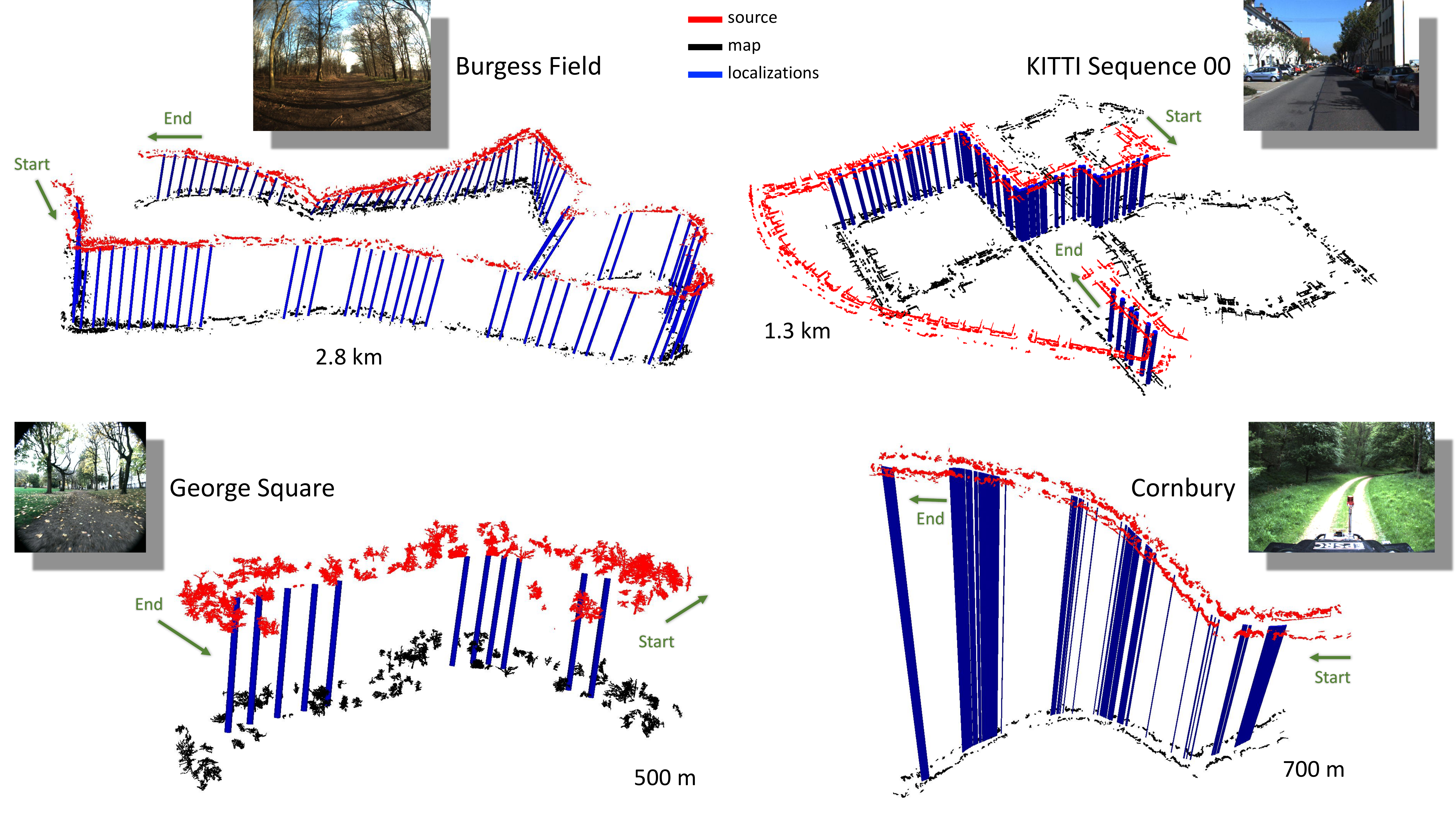}
	\caption{Qualitative visualization of the performance of the proposed approach (ESM) 
	on all datasets. The used datasets provide a variety of contexts to show the behaviour
	of ESM when coping with structured and unstructured locations.}
	\label{fig:combined_datasets}
\vspace{-3mm}
\end{figure*}

In the next set of experiments we aim to support our claim that the proposed 
algorithm performs comparably to the state of the art on both urban and natural 
environments when localizing, while requiring less computation. ~\tabref{tab:loc_results} 
shows the total number of loop closures detected in 
each of the datasets. For each algorithm we have optimized the parameters to 
retrieve the highest number of true localizations, while having zero false 
localizations. The learning approaches have not been retrained for each 
individual natural dataset but still manage to detect about two times more 
loop closures than the engineered methods. In brief, our algorithm has similar 
accuracy to~\cite{segmap2018} on all datasets. ~\figref{fig:bf_localizations} presents 
qualitative results for our approach compared to the baselines on the Burgess 
Field dataset, while also highlighting the computational times for each approach. 
In this environment the vehicle regularly detected loop closures during more 
than $2$\,km of operation. ~\figref{fig:combined_datasets} demonstrates the performance of our 
algorithm across all the datasets. The method did not produce any false 
localizations, while operating in real time on a CPU. 

\subsection{Computation Efficiency}
Finally, we evaluate the computational performance of the various 
approaches. We tested our approach on a mobile Intel Xeon E3-1535M CPU with
the following configurations: (1) single-core of the processor, (2) multi-core 
on the same processor, (3) CPU + NVIDIA Titan Xp GPU.
We have processed the push-broom point cloud data from Burgess Field and recorded 
the mean computational speeds. The target map 
consisted of $2783$ segments --- individual trees, bushes, and interleaved 
shrubs. Individual source clouds were created for every $22$\,m travelled with 
each source cloud containing an average of $46$ segments. In order to provide a 
fair 
comparison, we have utilized the same Euclidean segmentation method across 
algorithms, to create the same segments. This is important for the 
description stage of each pipeline. The segmentation was carefully parametrized 
to 
increase localization 
performance in this challenging natural environment. To estimate the pose we have 
used the same geometric consistency algorithm for all baselines substituting 
RANSAC~\cite{fischler1981random} 
with PROSAC~\cite{prosac2005} for the proposed method to incorporate 
the quality of the features. In this setting, using an incremental geometric 
consistency as in~\cite{segmap2018} is not possible, as 
the push-broom LIDAR provides only a single observation of a segment.

We have empirically evaluated the number of neighbors in feature space ($K$) on 
Burgess Field to retrieve the 
most positive localizations, while keeping zero false positives. For this 
experiment $K$ influences the KNN retrieval and pruning speeds. 
$K=25$ worked well for learning approaches as the features were very 
descriptive, 
while we kept $K=200$ for NSM and SegMatch.
~\tabref{tab:performance} summarizes the average runtime performance, per point 
cloud, in milliseconds. The size of the descriptor dimension corresponds directly 
to the computation time taken to describe a segment. The preprocessing and 
descriptor times scale linearly with the 
number of 
segments in a live point cloud ($\approx46$ in Burgess Field). The time 
required for matching also depends on the embedding dimension. Point clouds tend 
to be of similar size, thus the computation time for the 
segmentation does not vary. The pose estimation depends on the number of segments 
and their closest neighbors ($K$) and the pose estimator approach.
The total size of the SegMatch descriptor is 647 
dimensions, only 7 of which are used during the matching stage. These 647 are 
compressed efficiently to the 45 processed by the Random Forest. This results in 
fast KNN retrieval and slower RF pruning. The total size of 
the NSM descriptor is 66, which are extended to 330 for the pruning stage, making 
it less efficient for KNN retrieval and pruning. The SegMap descriptor requires 
PCA alignment and voxelization of the segments as preprocessing steps, after which 
the forward pass of the model is 
executed in C++. Even though the implementation is efficient, the $9.3$\,M 
parameters of the network describe a single segment in $0.5$\,s on a 
CPU.~\tabref{tab:performance} also summarizes the recorded GPU times for the 
learned approaches with $K=25$ during matching.

We focused our analysis on CPU-only solutions due to the particular efficiency 
of our approach. Our network does not require expensive pre-processing of the 
segments, the model consists of 4 $X$-conv layers with an input of 256 points. 
This results in only $300$\,K parameters, which represents the number of 
operations needed to be performed. Compared to other learning 
methods our network has $30$--$65$x less parameters, which allows the model to 
work in real time on a CPU. In addition, the embedding dimension is 
kept to just 16 dimensions which speeds up matching.

\section{Conclusion}
\label{sec:conclusion}

In this paper, we presented a novel descriptor for place recognition based on 
LIDAR segment matching in both urban and natural environments.
Our method exploits an efficient deep learning architecture that operates 
directly on point cloud data without the need of extensive preprocessing. This 
allows us to successfully detect loop closures at $\approx1$\,Hz while using only 
a CPU. 
We implemented and evaluated our approach on four different datasets containing 
natural forest and parkland as well as urban scenes and provided comparisons to 
other existing techniques. Our approach operates in real time on a CPU and 
achieves performance comparable to the state of the art, 
SegMap~\cite{segmap2018}, which requires a GPU to run in real time. The 
experiments suggest that our approach can be applied to mobile robots with 
limited computational power. In future work we are interested in deploying the 
approach on an ANYMAL quadruped and UAVs both of which lack a GPU.

\section{Acknowledgements}
We would like to thank Oliver Bartlett and our colleagues at the Oxford Robotics 
Institute for the datasets.

\bibliographystyle{IEEEtran}
\bibliography{ra-l}

\end{document}